\pdfoutput=1
\documentclass[11pt]{article}

\usepackage{EMNLP2022}
\usepackage{times}
\usepackage{latexsym}

\usepackage[T1]{fontenc}
\usepackage[utf8]{inputenc}

\usepackage{microtype}

\usepackage{inconsolata}

\usepackage{todonotes}
\usepackage{tabularx}
\newcolumntype{P}[1]{>{\centering\arraybackslash}p{#1}}
\newcolumntype{Y}{>{\centering\arraybackslash}X}

\title{A Report on the Euphemisms Detection Shared Task}

\author{Patrick Lee  \and Anna Feldman \and Jing Peng\\
      Montclair State University\\New Jersey, USA\\
      \texttt{\{leep6,feldmana,pengj\}@montclair.edu}\
      \\}
\begin{document}
\maketitle
\begin{abstract}
This paper presents The Shared Task on Euphemism Detection for the Third Workshop on Figurative Language Processing (FigLang 2022) held in conjunction with EMNLP 2022. Participants were invited to investigate the euphemism detection task: given input text, identify whether it contains a euphemism. The input data is a corpus of sentences containing potentially euphemistic terms (PETs)  collected from the GloWbE corpus \cite{davies2015expanding}, and are human-annotated as containing either a euphemistic or literal usage of a PET. In this paper, we present the results and analyze the common themes, methods and findings of the participating teams.
\end{abstract}

\section{Introduction}

Euphemisms are mild or indirect expressions that are used in place of other ones when discussing potentially offensive or sensitive topics. Their linguistic functions include (politeness, concealment, and neutralization of unappealing words/phrases). Despite being an important element of language use, their figurative nature poses a challenge for natural language processing (NLP).

There are numerous challenges to working with euphemisms. One is the phenomenon of the “euphemism treadmill”, by which words/phrases gain or lose euphemistic meanings over time \cite{pinker2003blank}. Another is that researchers may not agree on what euphemisms are. For example, \citet{zhu2021euphemistic,zhu2021self} treat code words as euphemisms, but our working definition does not.
%Some refer to “euphemisms” as including dysphemisms (e.g., x-phemisms in \citet{felt2020recognizing}) or code-words \cite{zhu2021euphemistic,zhu2021self}, w hile our working definition excludes these phenomena\todo{Is it what you wanted to say?}. 
Even when restricted to our working definition, however, annotators were found to sometimes disagree in the task of labeling example sentences as euphemistic \cite{gavidia-etal:2022}.  For all these reasons, the words/phrases in this shared task are referred to as potentially euphemistic terms (PETs).
The main challenge, which is the focus of this shared task, is the ambiguity of PETs: the same words/phrases that may be euphemistic in one context may be literal in another. For example, 

\noindent
\newline
\emph{Asked to choose \underline{between jobs} and the environment, a majority – at least in our warped, first-past-the-post system – will pick jobs.} (non-euphemistic)

\noindent
\newline
\emph{This summer, the budding talent agent was \underline{between jobs} and free to babysit pretty much any time.} (euphemistic)
\newline

We propose the Shared Task on Euphemism Detection: given input text, identify whether it contains a euphemism; i.e., distinguish between euphemistic and literal usages of the same PETs in different contexts. The data used is a corpus of texts containing PETs, collected by \citet{gavidia-etal:2022}, which contains parallel euphemistic and literal examples for a range of PETs. 46 participants spanning 13 teams attempted the task, and we received 9 system descriptions. 

Due to a lack of extensive research in this area, it is unclear how NLP techniques, such as language models (LMs), will perform on euphemism detection. The purpose of this shared task is to (1) explore the ability of NLP techniques for this task and (2) investigate what methods could further improve upon their performance. 

\section{Related Work}

There is not much work on automatic detection of euphemisms. 
The most directly related work is by \citet{magu2018determining}, \citet{felt2020recognizing}, \citet{kapron2021diachronic}, \citet{zhu2021self} and \citet{zhu2021euphemistic}.
\citet{felt2020recognizing} present the first effort to recognize  euphemisms and dysphemisms (derogatory terms) using NLP. The authors use the term  \emph{x-phemisms} to refer to both. They first identify three sensitive topics (lying, stealing, and firing). They use a weakly supervised algorithm for semantic lexicon induction \cite{thelen2002bootstrapping} to generate lists of near-synonym phrases for each topic semi-automatically. \citet{felt2020recognizing} experiment with two methods to classify phrases as euphemistic, dysphemistic, and neutral: 1)  dictionary-based method addressing affect, connotation, intensity, arousal, and dominance; 2) contextual sentiment analysis to classify x-phemisms. The important product of this work is a gold-standard dataset of human x-phemism judgements. The important lesson here is that \citet{felt2020recognizing} show that sentiment connotation and affective polarity are useful for identifying x-phemisms, but not sufficient. While the performance of \citet{felt2020recognizing}'s system is relatively low and the range of topics is very narrow, this work certainly inspires further investigations. 

\citet{zhu2021self} define two tasks:  1) euphemism detection (based on the input keywords, produce a list of candidate euphemisms) 2) euphemism identification (take the list of candidate euphemisms produced in (1) and output an interpretation). \citet{zhu2021self} select sentences matched by a list of keywords, create masked sentences (mask the keywords in the sentences) and apply the masked language model proposed in BERT \cite{devlin2018bert} to filter out generic (uninformative) sentences and then generate expressions to fill in the blank. These expressions are ranked by relevance to the target topic. %In other words, they use the masked language model twice: 1) to filter the masked sentences and 2) generate the euphemism candidates from the masked sentences. 

%\cite{kapron2021diachronic} investigate gender differences in language and the assumption that women use euphemisms more than men. They also emphasize that euphemisms drive language change and hypothesize that if women use more euphemisms, then women also contribute to language change more \cite{labov2002driving}. \cite{kapron2021diachronic} use four large diachronic text corpora of English to evaluate the claim that women
%use euphemisms more than men through a
%quantitative analysis. They assemble a list of
%106 euphemism-taboo pairs to analyze their
%relative use through time by each gender in the
%corpora. Contrary to the existing belief, \cite{kapron2021diachronic}'s %results show that women do not use euphemisms
%with a higher proportion than men. 

%\cite{xu2021blow} propose a new dataset of Chinese \emph{cants}, also known as doublespeak, cryptolect, argot, anti-language, or secrete language. \emph{Cant} is the jargon or language of a group, often employed to exclude or mislead people outside the group.  \cite{xu2021blow} formulate a task for cant understanding and provide both quantitative and qualitative analysis for tested word embedding similarity and pretrained language models. Their experiments show a great gap between human performance and model results indicating that the task is challenging. 

Euphemisms are also related to the language of politeness (e.g.,  \citet{danescu2013computational,rababah2014translatability}), which plays a role in applications involving dialogue and social interactions in different contexts. 

Other shared tasks have proposed a similar classification task on other types of figurative language. \citet{sarcasm-detection-task:22} report on a sarcasm detection task run on conversation data from Twitter and Reddit, while \citet{idiom-detection-task:22} report on an idiom detection and embedding task. 

\section{Task Setting}

Participants were given a dataset of PET-containing texts created by \citet{gavidia-etal:2022}. In this section, we describe the dataset and the classification task.

\subsection{Dataset Description}

The corpus of PETs was formed by taking a list of PETs (single and multi-word expressions, collected from a variety of sources) and extracting texts from the GloWBe corpus \cite{davies2015expanding}  (only the US-English portion) which contained them. Each text sample comprised up to 3 sentences: the sentence that the PET appeared in, as well as the preceding and following sentences, if available. In total, the dataset contains 1,965 text samples spanning 129 different PETs and 7 topics/categories. Of these, 1,382 were annotated to contain a euphemistic usage of a PET, and the remaining 583 a literal usage. Thus, the dataset is imbalanced (an aspect which multiple teams explicitly considered in their approaches). The full details of the data, including the distribution amongst the PETs and topics, can be found in the original paper \cite{gavidia-etal:2022}. 

The training and test sets were created using an 80-20 split. The range of PETs which appeared in each split was balanced as much as possible, given that several PETs only appeared once as a euphemistic or literal example. Details of the split are summarized in Table \ref{tbl:Dataset}.

\begin{table}[htb]
\begin{center}
\begin{tabularx} {\columnwidth}{|Y|Y|Y|Y|Y|Y|}
\hline
\textbf{Split} & \textbf{Rows} &  \multicolumn{2}{c|}{\textbf{Euphemistic}} & \multicolumn{2}{c|}{\textbf{Literal}}  \\ 
\cline{3-6}
 & & Rows & PETs & Rows & PETs \\
\hline
Train & 1572 & 1106 & 122 & 466 & 54 \\
\hline
Test & 393 & 276 & 121 & 117 & 55 \\
\hline
\end{tabularx}
\end{center}
\caption{Train-Test Split Details}
\label{tbl:Dataset}
\end{table}

A simplified version of the dataset was created for the participants, where each row contained only (1) the text sample with the PET denoted in brackets, and (2) its label (a '1’ for euphemistic, and a ‘0’ for literal). This version omitted information about each row, such as meta-information about the PET: the specific morphological variant present in the row, the topic category (e.g., "death", "politics", etc.), and whether it always appeared in the dataset as a euphemism (“always-euphs”) or only sometimes (“sometimes-euphs”). This information, however, was available to the participants via a Github link to the full dataset (which several teams chose to leverage).

\subsection{Task Description}

The shared task was set up as a competition on Codalab\footnote{\url{https://codalab.lisn.upsaclay.fr/competitions/5726}}. Participants were invited to develop systems trained on the training data (see Table \ref{tbl:Examples} for some examples) and submit answer labels on the test set, which would be compared to the labels in the original dataset for evaluation. The evaluation metric used to rank submissions was the macro-F1 score. 

\begin{table}[htb]
\begin{center}
\begin{tabularx}{\columnwidth}{|X|c|}
\hline
\textbf{Text} & \textbf{Label} \\
\hline
More likely it'll harm them. With less products to make, Foxxcom will have to <lay off> workers. With more workers seeking jobs those other factories will be able to resist demands for higher wages. & 1 \\
\hline
We do NOT need some self-imposed book cop telling us what to read or not to read. <Lay off>. Get over it. & 0 \\
\hline
After about 30 minutes of waiting, a fight broke out between an older African American man and an African American woman of <a certain age>. After making a lot of noise and landing a few blows to their respective bodies, the armed security guards escorted them out of the terminal. & 1\\
\hline
\end{tabularx}
\end{center}
\caption{Example Rows from the Dataset}
\label{tbl:Examples}
\end{table}

% \multicolumn{1}{l|}{weed}& \textit{euphemistic} & You will want to stop off at the   medical marijuana dispensary for a supply of fireworks, alcohol, personal   weaponry and dope-\textless weed\textgreater. Then, fill a glass or pop a top   or load a bong or whatever one does, to get along these days.\\
% \multicolumn{1}{l|}{}& \textit{literal}& In some ways, cultivating for   \textless weed \textgreater control is almost a lost art. Herbicides seemed   to work so well for so long that many farmers abandoned mechanical means of   control.\\ \hline
% \multicolumn{1}{l|}{disabled} & \textit{euphemistic} & No no no no. I'm in the same   situation-- \textless disabled \textgreater, chronic pain, artist, no   "visible disability" (even when I'm in my chair), and nobody   understands that it takes us longer to do *everything*. I'm honestly   surprised you even humored your neighbor this far!\\
% \multicolumn{1}{l|}{}& \textit{literal}& They claim there is no network   or storage capability in these machines, clearly this is not true. These   features may be \textless disabled \textgreater or only available to   administrators who service the equipment, but in any event the TSA @ @ @ @ @   @ @ @ @ @ problems. As to the veterans out there who work for the TSA, I   share your frustration \\

\section{Participants and Results}

8 teams that participated in the task also submitted descriptions of their systems, with one author additionally exploring a zero-shot/few-shot variant of the task. A summary of their performances is shown in Table \ref{tbl:Results}. In this section, we describe the methods used by the best-performing teams, and analyze the common themes between the approaches and motivations of all the submissions. 

\begin{table*}[!h]
\begin{center}
\begin{tabularx}{\textwidth}{|c|c|c|Y|} 
\hline
\textbf{Rank} & \textbf{Username} & \textbf{Macro-F1} & \textbf{Title of Paper} \\
\hline
1 & vgangal & 0.88 & EUREKA: Euphemism Recognition Enhanced Through KNN-based Methods and Augmentation \cite{eureka:22} \\
\hline
2 & ilker & 0.87 & Detecting Euphemisms with Literal Descriptions and Visual Imagery \cite{kesen-etal:22}\\ 
\hline
3 & Wanderer & 0.85 & A Prompt Based Approach for Euphemism Detection \cite{abu:22}\\
\hline
4 & liuyiyi & 0.84 & Euphemism Detection by Transformers and Relational Graph Attention Network \cite{wang-etal:22}\\
\hline
5 & peratham.bkk & 0.82 & TEDB System Description to a Shared Task on Euphemism Detection 2022 \cite{peratham:22}\\
\hline
6 & PaulTrust & 0.79 & Bayes at FigLang 2022 Euphemism Detection shared task: Cost-Sensitive Bayesian Fine-tuning and Venn-Abers [...] \cite{trust:22}\\
\hline
7 & devika & 0.74 & An Exploration of Linguistically-Driven and Transfer Learning Methods for Euphemism Detection \cite{tiwari:22} \\
\hline
8 & gunetsk99 & 0.72 & Adversarial Perturbations Augmented Language Models for Euphemism Identification \cite{guneet:22}\\
\hline
\end{tabularx}
\end{center}
\caption{Results of submitted systems to the Shared Task on Euphemism Detection}
\label{tbl:Results}
\end{table*}

\subsection{Best Submissions}

The best-performing team \cite{eureka:22} (macro-F1 0.881) explores a variety of data and modeling modifications, and combine the best-performing ones into an ensemble of three models to improve upon a baseline RoBERTa-large model \cite{liu2019roberta}. On the data side, they explore two methods of data augmentation, and find that adding examples containing similar/opposite word senses to PETs (for positive and negative examples) works best; this highlights the potential significance of sense-based approaches for this task. They also explore identifying and correcting 25 potentially mislabeled rows from the dataset, reporting an improvement of 0.0036 points over the base model and a final score \(\sim \)0.007 higher when using their “cleaned” dataset. While it is unclear how this cleaning would affect other systems in general, their investigation hints at the potential for not only human disagreement but also human error in labeling figurative language. On the modeling side, they find that classifying on the tokens of the PETs, rather than the [CLS] token, yields significant a performance increase. They also experiment with two methods of incorporating extra context and find that k-Nearest Neighbor (kNN) (5NN in this case) augmentation yields slight improvements. They report the best improvement by combining the following three models: (1) a RoBERTa-large model classifying on the PET token(s), (2) the same, but using their sense-augmented dataset, and (3) the same as (1), but interpolating the classification probabilities of its base model and the 5NN classifier.

The second-best performing team \cite{kesen-etal:22} achieved a macro-F1 of 0.872 using additional supervision and, interestingly, incorporating visual imagery into their approach. Using DeBERTa-v3 \cite{he2021debertav3} as their baseline (the “large” version of which performs best), they incorporate additional supervision by including PETs themselves, as well as their (manually collected) literal descriptions, in their inputs. The authors found this to greatly improve performance, reasoning that such direct supervision could help mitigate ambiguity inherent to the task. This is a similar result to  \cite{eureka:22}, where extra attention on the PETs themselves is effective. They then obtain imageries of both the PETs and their literal descriptions using a text-to-image model, and obtain image embeddings using a pretrained visual encoder. These are incorporated into training, and yields statistically significant performance improvements. Qualitative analysis of the images for each PET reveals insights into how LMs might understand figurative expressions.

\subsection{Analysis of Methods}

Below, we describe approaches that we observed in multiple submissions. Since the objective was to explore different aspects of this task, we find these insights to be valuable, even if they did not score high.

\subsubsection{Using PET Embeddings Directly}

Multiple teams found that explicitly involving the tokens of the PET for classification led to significant improvement. \citet{kesen-etal:22} and \citet{wang-etal:22} include the PET in their inputs by concatenating it to each input text prior to learning. \citet{eureka:22} run their final classification on the PET embedding, rather than that of the [CLS] token. These changes were a significant feature of these teams’ best approaches. It appears that providing direct supervision/focusing the modeling procedure on the PET helps, perhaps because the PET is the semantic focus of the task (rather than other words in the data, which are not always important).

\subsubsection{Using the Literal Meanings of PETs}

Each PET is detailed in \citet{gavidia-etal:2022} to have a literal meaning or paraphrase, which is the more offensive or unpleasant “real meaning” that the PET substitutes. Several teams chose to integrate literal meanings of each input PET into their methods, though they opted to generate their own literal meanings, rather than use the ones from the original paper. This seemed to be effective, as the two best-performing teams found it to improve performance — \citet{kesen-etal:22} appended literal meanings to their inputs and used them to generate image embeddings, while \citet{eureka:22} used literal meanings to select examples for data augmentation — in conjunction with direct supervision on the PET (4.2.1). \citet{tiwari:22} paraphrased PETs with their literal meanings in attempt to obtain sentiment shifts, but this did not work well for classification, likely because the paraphrasing mechanism did not produce quality paraphrases that could serve as literal meanings for the PETs.

\subsubsection{Addressing Data Imbalance/Inadequacy}

Multiple teams addressed the fact that the dataset was imbalanced (see \ref{tbl:Dataset}). \citet{abu:22} found that their approach scored an F1 of 0.914 versus 0.789 on the euphemistic and literal examples, respectively, suggesting that the model performance could indeed be skewed towards the higher-volume euphemistic examples. 

\cite{trust:22} experimented with multiple modeling enhancements, such as Bayesian modeling, exclusively to address the imbalance issue and found that they improved performance. \citet{guneet:22} sought to address the imbalance by using adversarial perturbations to augment the ‘0’ label, and found it to increase performance slightly. As aforementioned, \citet{eureka:22} augment the dataset by strategically selecting sentences from other corpora (albeit for general augmentation, rather than to achieve a balance) and similarly report slight performance increases, but, like \citet{guneet:22}, they do not exclusively use the augmented data in their final approach. These results generally support the intuition that addressing data inadequacies helps models learn, but only partially for this shared task.

\subsubsection{Incorporating Additional Context}

While this task necessitates making use of contextual differences between input texts, several teams attempted to incorporate additional information from the context. \citet{wang-etal:22} model syntactic connections between the PET and other words in the text as a relational graph, finding that using this as an input to BERT is effective (though no baseline or example parse is provided). As aforementioned in Section 4.1, \citet{eureka:22} slightly improve performance by using the k-nearest neighbors for each input as additional context. \citet{guneet:22} finds that simply using longer input sequence lengths than standard BERT allows for (512) generally improves performance, although it is not clear at what sequence length this would cease to be the case. All these methods to incorporate more input context seemed to generally improve performance.

\subsubsection{Linguistic Transparency}

Some teams attempted solutions that would promote model explainability. \citet{kesen-etal:22} claim that images of PETs and their literal meanings are a way to gain insight into how models interpret PETs, but while this is an interesting way to probe models' current understanding of figurative expressions, it is unclear how models might use these images to enhance classification in an understandable way. The two-model ensemble used by \citet{wang-etal:22} attempt to incorporate linguistic (semantic and syntactic) information of the PET and its context, though without compelling examples of how these may help classification, the improvements seem somewhat unexplainable. Finally, \citet{tiwari:22} interestingly pursued an approach that is based exclusively on the linguistic intuition that euphemisms produce sentiment shifts, but using these shifts alone for this task was ineffective.

On the other hand, methods which found success using transformers are not very transparent. \citet{peratham:22} test various transformers and obtain their best result by combining a CNN variant with the highest-performing one, but it is not clear what this network is learning, as is typically the case with neural networks. Furthermore, \citet{keh:22} find that transformers work decently for this task in the zero-shot setting (see 4.2.6), but admit that it’s not clear what BERT is learning in order to do so.

\subsubsection{Zero/Few-shot Learning}

\citet{abu:22} train a RoBERTa model using prompt-tuning because it has been shown to work well (better than regular fine-tuning) with few-shot examples. While our task was not formulated as a zero/few-shot task, several PETs appeared only a few times in the data and were effectively few-shot examples. 
\citet{keh:22} notably re-formulate the task for the zero/few-shot setting. When PETs were randomly selected to be zero-shot examples, RoBERTa-large achieved a score of 0.740, showing that the model was able to “learn” something about euphemisms (not simply memorizing) and apply it to examples with PETs unseen during training. They also show that few-shot examples benefit the model greatly, with 3-shot performance (0.825) nearly matching the baseline performance (0.836). Furthermore, they found that GPT3, which typically works well in the zero/few-shot setting, worked badly for this task.

\section{Discussion and Findings}

Here, we describe some common findings of the submitted systems that may be useful for future work.  

% We also discuss the significance of this shared task for figurative language understanding in general.

% \begin{enumerate}

% \item \textbf{More data is better.} 

\subsection{More Data is Better}

Having more examples of each PET generally led to better performance. \citet{kesen-etal:22} and \citet{eureka:22} improved performance by augmenting the dataset. \citet{abu:22} showed that performance on the euphemistic label is better, likely because there were many more examples than the non-euphemistic label. Compellingly, \citet{keh:22} showed in their zero/few-shot task that 3-shot learning was much better than 1-shot. The results of this task call for larger datasets of euphemisms, perhaps from a variety of different sources.

\subsection{BERT Works} 
All teams experimented with some variation of BERT and reported decent scores using unmodified BERT models, the highest being 0.839 \cite{kesen-etal:22} using RoBERTa-large. The zero-shot investigation by \citet{keh:22}, too, shows that RoBERTa picks up something during training that is generalizable to other euphemisms. Overall, pre-trained transformers seem to provide a solid baseline from which to launch euphemism work. 

\subsection{Linguistic Intuitions}
%\item \textbf{Linguistic intuitions of euphemisms were correlated with helpful methods}.
 
Because euphemisms (as well as other types of figurative language) are often commonly used expressions, it is likely that large language models have some existing “knowledge” about these collocations. One could interpret the success of using PET embeddings directly \cite{kesen-etal:22,wang-etal:22} as evidence that models can leverage this knowledge for the task. 

Another linguistic notion is that euphemisms’ function may lead to changes in sentiment, which has been found to potentially be useful for identifying euphemisms \cite{felt2020recognizing,lee2022searching}, but it remains somewhat unclear whether it is useful for the disambiguation proposed in this task. \citet{peratham:22} do find that transformers pre-trained specifically on sentiment were more helpful than those pre-trained on other tasks (e.g., sarcasm or hate speech detection). \citet{tiwari:22} try a non-transformer-based approach based on the intuition that PETs should produce higher sentiment shifts in euphemistic sentences when paraphrased with its literal meaning, but found it was difficult to generate such paraphrases. This corroborates our own past experimentation \cite{lee2022searching}, and it seems that future approaches based on sentiment shifts have to address the need for better paraphrasing mechanisms, or consider using them to supplement a larger input feature set.
% \end{enumerate}

\section{Conclusions and Future Work}

We present the results of “The Shared Task on Euphemism Detection for the Third Workshop on Figurative Language Processing” and summarize the various systems submitted, as well as common findings. Overall, we find that results are promising: even when dealing with the difficult issue of an especially polite and indirect form of figurative language, current NLP techniques such as transformers and augmentation seem to work quite well. Teams explored a variety of intriguing methods to enhance the baseline performance of these models, some of which were even linguistically transparent. If one considers that labeling euphemisms is subject to human disagreement, the F1-scores achieved by the teams are even more compelling since they may be near, if not already at, the level of human agreement on the task. The results of this shared task establish a baseline for future work on euphemisms and figurative language in general. 

Future work on this task could be expanding on the dataset to include more examples and a wider range of PETs, testing further enhancements, and improving performance by ensembling various combinations of the best-performing improvements. Future work for euphemism detection in general could be to expand from the disambiguation task; e.g. identifying where euphemisms are in a text, providing interpretations of a euphemism, or even euphemistic language generation.  

%\section{Bib\TeX{} Files}
%\label{sec:bibtex}
%Unicode cannot be used in Bib\TeX{} entries, and some ways of typing special characters can disrupt Bib\TeX's alphabetization. The recommended way of typing special characters is shown in Table~\ref{tab:accents}.

%Please ensure that Bib\TeX{} records contain DOIs or URLs when possible, and for all the ACL materials that you reference.
%Use the \verb|doi| field for DOIs and the \verb|url| field for URLs.
%If a Bib\TeX{} entry has a URL or DOI field, the paper title in the references section will appear as a hyperlink to the paper, using the hyperref \LaTeX{} package.

\section*{Limitations}
While the data used denoted where the target PET was in each text sample, this information is not provided in raw text. Identifying the PET in a text sample is a challenge that future approaches, especially those seeking to focus models on the PET, will need to consider. Additionally, this shared task was run on a dataset that could be significantly expanded and balanced. The dataset also contained potentially subjective labels that were only made by two human annotators; this could be made more robust by ensembling more annotators. Finally, this task was based on a dataset comprising only of texts of US English, and it is unclear how these results would transfer cross-lingually to other kinds of euphemisms.

\section*{Ethics Statement}
When we created the shared task, we tried to be compliant with the \href{https://www.aclweb.org/portal/content/acl-code-ethics}{ACL Ethics Policy}. Euphemisms are expressions that `hide' prejudices by using softened language. Models capable of recognizing and interpreting euphemisms should be better at detecting biases related to gender, age, race, or socioeconomic background, detrimental to the society.

%We encourage all authors to include an explicit ethics statement on the broader impact of the work, or other ethical considerations after the conclusion but before the references. The ethics statement will not count toward the page limit (8 pages for long, 4 pages for short papers).

\section*{Acknowledgements}
This material is based upon work supported by the National
Science Foundation under Grants No. 1704113 and  No. 2226006.

\bibliography{anthology,custom}
\bibliographystyle{acl_natbib}

%\appendix

%\section{Example Appendix}
%\label{sec:appendix}

%This is a section in the appendix.

\end{document}